\documentclass{article}
\usepackage[utf8]{inputenc}
\usepackage{booktabs} 
\usepackage{amsfonts}
\usepackage{amssymb}
\usepackage{wasysym}
\usepackage{graphicx}
\usepackage{epstopdf}
\usepackage{hyperref}
\usepackage{authblk}
\usepackage [english]{babel}
\usepackage [autostyle, english = american]{csquotes}
\pdfoutput=1
\MakeOuterQuote{"}

\title{The Diagrammatic AI Language (DIAL): Version 0.1}

\author{Guy Marshall}
\author{Andr\'e Freitas}

\affil{School of Computer Science\\ University of Manchester\\ 
United Kingdom}

\date{December 2018}

\begin{document}

\maketitle


\begin{abstract}
    Currently, there is no consistent model for visually or formally representing the architecture of AI systems. This lack of representation brings interpretability, correctness and completeness challenges in the description of existing models and systems. DIAL (The Diagrammatic AI Language) has been created with the aspiration of being an "engineering schematic" for AI Systems. It is presented here as a starting point for a community dialogue towards a common diagrammatic language for AI Systems. 
\end{abstract}

\section{Introduction}
\label{section:motivation}
\subsection{Motivation}

The development of real-world Artificial Intelligence systems requires the complex coordination of multiple components. Currently, there is no consistent model for visually or formally representing the architecture of AI systems. This lack of representation brings interpretability, correctness and completeness challenges in the description of existing models and systems. 

In the context of scientific communication, most approaches and systems today are described by a combination of arbitrary diagrammatic elements, algorithms, formulae and natural language descriptions. In this paradigm, there is little consistency on abstraction levels (for example, it is common for diagrams to refer to formulae terms) or notation. 

From the perspective of scientific practice, these limitations challenge dialogue, transparency and reproducibility. This work outlines the initial specification, design and application of a diagrammatic language for describing complex, multi-component AI systems. The goal of the language, named `DIAL’ (Diagrammatic AI Language), is to reduce the communication effort between AI researchers and engineers for describing, interpreting and reasoning over complex AI systems.

\subsection{Background}

The idea of applying diagrammatic representations to AI is not new. In 1985, Aaron Sloman writes about the need for representational formalisms in AI, praising diagrams for controlling the search space \cite{Sloman1984WhyFormalisms}. In a related way, the Sapir-Wharf hypothesis states that the structure of native language determines the native speaker's perception and categorisation experience, a conjecture which could be extended to diagrammatic languages. Following this thinking, we argue that for AI researchers and engineers, diagrammatic structures can determine perception and categorisation of AI systems and subsystems. As such, a diagrammatic language (such as DIAL) may aid in reasoning about AI systems.  

In the remainder of Section \ref{section:motivation} we will look at the main structured alternative diagrammatic approach (UML) and show the current heterogeneity issue with AI System diagrams. This will provide the necessary grounding to propose requirements for AI System diagrams, and introduce DIAL as a potential solution to these issues. The main diagrammatic components of DIAL are then described, formalised and exemplified. As this document focuses on the specification of DIAL v. 0.1, we concentrate on the narrative around the requirements and design for DIAL, seeking the engagement of a community around DIAL and the collection of critiques and contributions. 

\subsection{Key Benefits of DIAL}

The core benefits aimed for DIAL are:

\begin{enumerate}
\item Having a representation which supports AI systems design, reasoning and learning.
\item Increasing the consistency and efficiency of the communication around AI Systems. 
\end{enumerate}

The rapid evolution of techniques have been defining an increasingly dynamic and fragmented AI landscape. The plethora of proposed models and tools and their associated complexity brings an increasing difficulty to interpret, contrast, compare and select models in a principled and critical manner. DIAL is designed using a combined bottom-up/top-down design approach. In the bottom-up component, reoccurring features observed in existing diagrams are amalgamated into the representation, complemented by the translation of elements represented as text into the integrated diagrammatic representation. In the top-down component, high-level requirements are introduced, focusing on the completeness and interpretability aspects of the representation. 

DIAL aims at providing the means for a self-contained description of the core functional elements of an AI system from both research and engineering perspectives. Supporting datasets, data management infrastructures, gold standards and evaluation metrics are first-class citizens of the representation. Additionally, DIAL aims by design to provide a comprehensive set of components within the full AI spectrum, not privileging a particular category of techniques (e.g. Neural Networks).

The self-contained and concise aspects of DIAL can be used as a foundation for a possible way to materialise \textit{nano-publications} in the context of AI. Using DIAL, readers can navigate through a concise representation of an AI system or experiment, freeing the AI practitioner from the need to go through heterogeneous diagrammatic and textual descriptions. This format supports the ability to quickly understand novel research and approaches, and be refreshed of them quickly without having to reread a supporting textual description. In the short-term, it is hoped that the usage of DIAL will facilitate more effective consumption of AI System concepts.

\subsection{Why UML isn't a Good Answer}

Within software more broadly, the communication and representation of architectures is supported \cite{Shahin2014ATechniques} with standards such as UML \cite{Booch1998UnifiedThe} and DARWIN \cite{Harrison2005TheGlobalization}.

For more specific applications, different diagrammatic representations are also being created. For "user requirements specification", nine deficiencies in UML have been identified, including that "UML cannot model the behavior of high-level system components\ldots"  \cite{Glinz2000ProblemsLanguage}. In dynamic modeling, Otero \cite{Otero2005AnUML} found Open Modeling Language worked better than UML for both quality and comprehension ratings.

UML is also seen as overly complex, with UML Class Diagrams having a graphic complexity of 14 (where the human cognitive limit is six) \cite{Moody2007WhatDevelopment}. Perhaps the most compelling argument against the suitability of UML comes from a pragmatic space. The existence of UML is quite common knowledge, and yet is not being used by AI practitioners, perhaps for reasons described in the above paragraph.

Within AI, the increasing complexity of systems increases the associated effort in communicating and understanding the fundamental functional components of a system. In the context of AI research and practice, there is a natural entanglement between the scientific and engineering discourses. Additionally, the design of an AI system will involve the consideration of multiple abstraction levels, ranging from a formal mathematical description level, to their algorithmic realisation and upstream software materialisation. 

While this task imposes major challenges in a pure ML pipeline, with architectural elements, hyper-parameters, feature engineering and associated datasets, this effort grows exponentially in systems which involve the coordination of multiple and heterogeneous classification, analytic and inference steps. In this context, a diagrammatic representation for AI can provide an interpretation mechanism which captures the core methods, resources and interdependencies employed within the AI task in a holistic manner. 

\subsection{The Problem of Diagrammatic Heterogeneity}
\label{section:heterogeneity}

There is a large variety of AI Diagrams being used within the AI community at present. Taking the sub-area of Neural Networks as an example, a large variety of neural architectures have emerged recently. Some initiatives to collect and describe them systematically have surfaced \url{http://www.asimovinstitute.org/neural-network-zoo/}. However, as the author states: "One problem with drawing them [Neural Networks] as node maps: it doesn’t really show how they’re used". Tensorflow has graph visualisations (see \href{https://www.tensorflow.org/guide/graph_viz}{TensorBoard}) which are also instructive, at a lower level of abstraction, and tied to a specific technology.

In order to quantify this existing heterogeneity, we performed a short systematic literature analysis on the use of diagrams to represent NNs in the context of NLP systems. 

The analysis started by selecting 30 randomly sample papers from the overall 331 accepted papers from COLING 2018. For the sampled papers, diagrams of Neural Networks were manually extracted, and the diagrams categorised. Only the diagrams and the text or formulae included within those diagrams is analysed. The supporting context is purposefully ignored in order to examine the effectiveness of the diagrammatic representation itself. COLING 2018 was chosen exclusively in order to ensure the same social context. Prior to the study, we identified a set of core NN concepts often articulated through diagrammatic means, which have been captured in Table ~\ref{tab:NN_table}. 

Of the 30 papers selected at random, 23 (77\%) were manually classified as being supported by a NN-based classifier. Note that this was not filtered for Neural Network papers, and this perhaps serves to highlight the prevalence of this technique at this time.
Of the papers in some way concerned with Neural Networks, 78\% include a diagram of a Neural Network. Table ~\ref{tab:NN_table} shows the distribution of Neural Network diagrams with each conceptual property.

\begin{table}[ht]
    \caption{Neural Network Diagram Summary}
    \label{tab:NN_table}
    \vskip 0.15in
    \begin{center}
    \begin{small}
    \begin{sc}
    \centering
    \begin{tabular}{lr}
        \toprule
        Concept or Attribute & Frequency\\
        \midrule   
        Abstraction: High level system & 67\% \\
        Abstraction: Intermediate level & 50\%\\
        Abstraction: Component level & 28\% \\
        Abstraction: Mixed levels & 6\% \\
        Multiple diagrams & 44\% \\
        Processing Layer Organisation & 61\% \\
        Classifier Type & 33\% \\ 
        Equational or Mathematical Elements & 72\% \\
        Vector or Tensor Operation  &  33\% \\
        Example input, output or processing & 39\% \\ 
        Dimensionality & 11\% \\
        Language Elements  & 94\% \\
        Recurrent operations & 94\% \\
        \bottomrule
    \end{tabular}
    \end{sc}
\end{small}
\end{center}
\vskip -0.1in
\end{table}

The vast majority of diagrams are high level diagrams, complemented by textual labels (e.g. LSTM). However, given the diagrams from a paper at random, it is often impossible to deduce fundamental aspects of the system, such as the functional elements. Table ~\ref{tab:NN_table} shows that few diagrams include classifier types, nor dimensional or compositional information. 

DLPaper2Code categorises Deep Learning model diagrams into five categories: Neuron plot, 2D box, Stacked2D box, 3D box and Pipeline plot \cite{Sethi2018DLPaper2Code:Papers}. The authors showed this coarse-grained classification useful, though this obfuscates the cognitive complexity involved with the variety of representations within each category. There is a huge variety of semiotic principles utilised in different ways, from semiotic choices such as how colour is used, through to "icon" choices, and even in terms of semantic content (i.e. what they choose to represent). 

\begin{figure}[htbp]
    \centering
    \fbox{
    \includegraphics[scale=0.5]{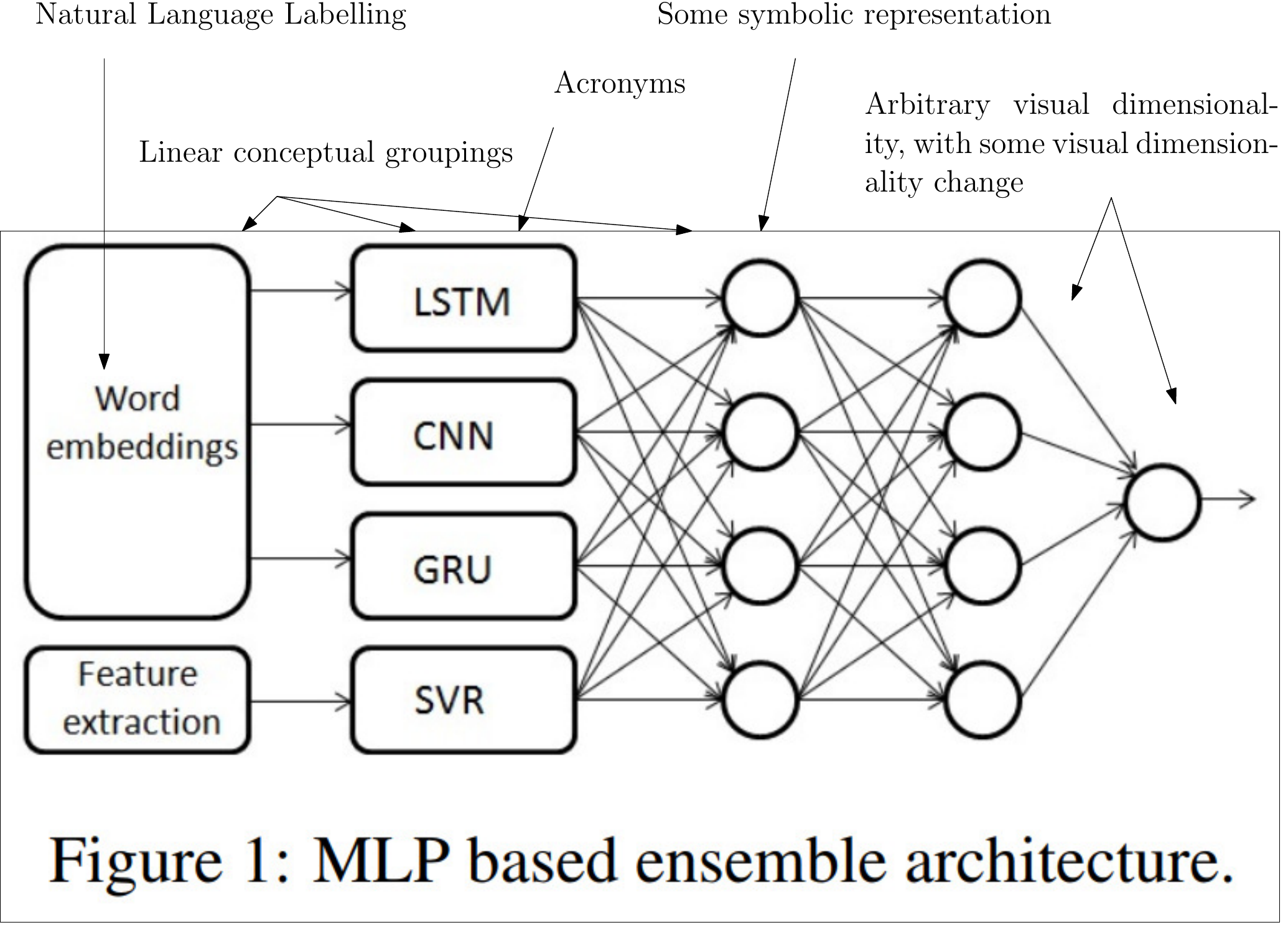}
    }
    \caption{Diagram for a Multilayer Perceptron Ensemble \cite{Akhtar2017AAnalysis}}
    \label{fig:akhtarExample}
\end{figure}

\begin{figure}[htbp]
    \centering
    \fbox{
    \includegraphics[scale=0.37]{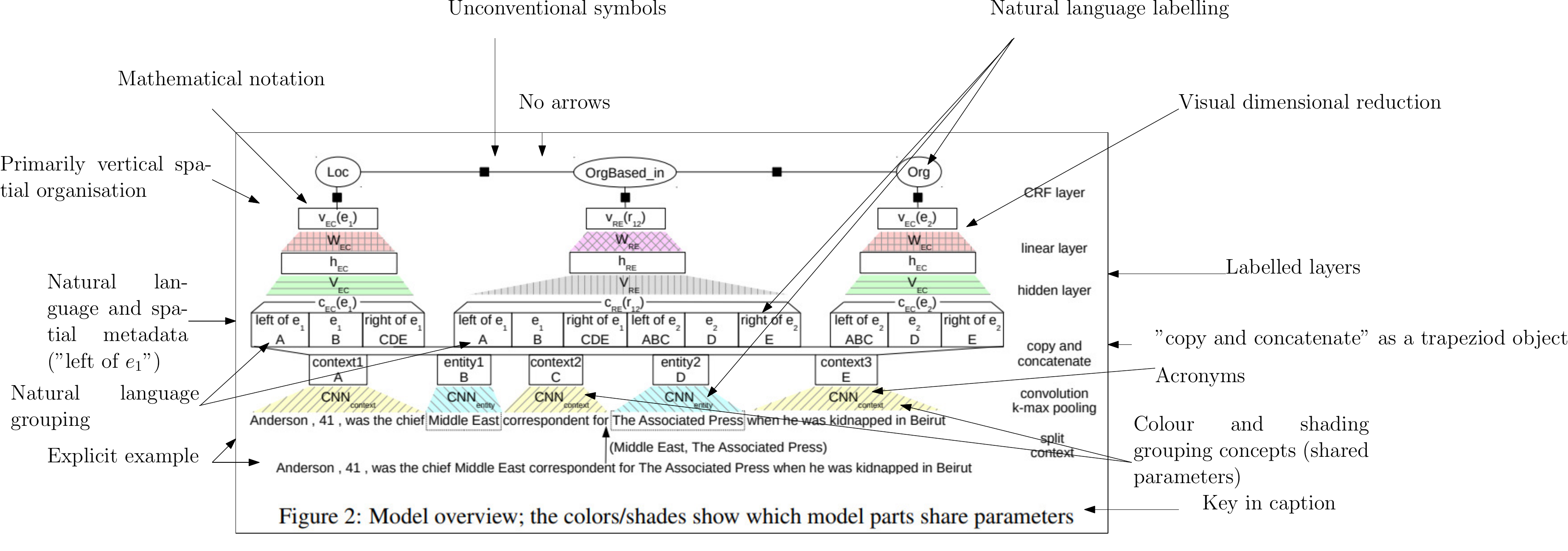}
    }
    \caption{Global Normalization of CNN \cite{Adel2017GlobalClassification}}
    \label{fig:adelExample}
\end{figure}

\begin{figure}[htbp]
    \centering
    \fbox{
    \includegraphics[scale=0.5]{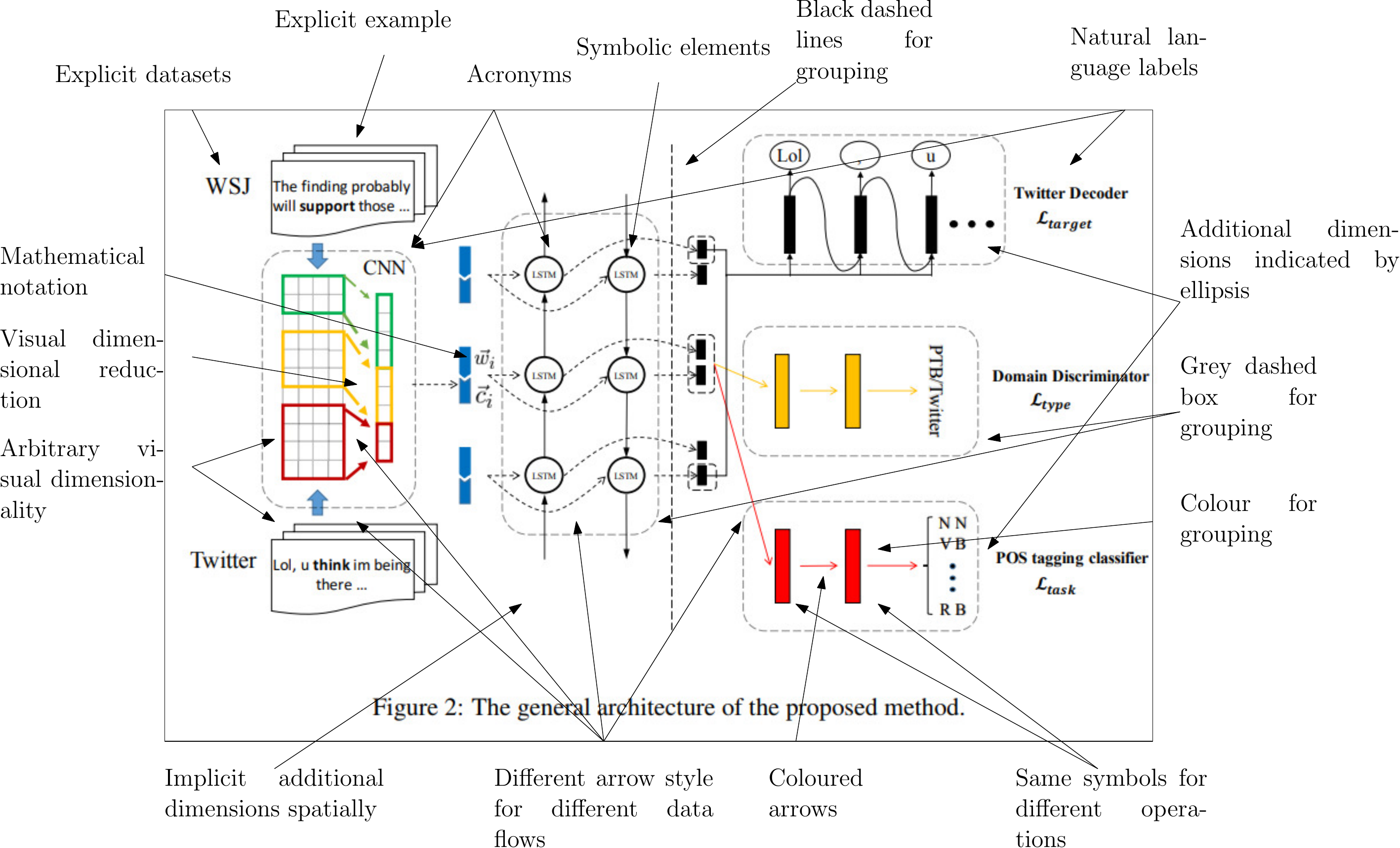}
    }
    \caption{Part-of-Speech tagging with adversarial neural networks \cite{Gui2017Part-of-SpeechNetworks}}
    \label{fig:guiExample}
\end{figure}

An effective way to informally see the heterogeneity issue is by observing concrete examples of diagrams in use. A simple randomised selection of papers from the main AI conferences can demonstrate this. In the following examination, we include a brief commentary on some diagrams presented as part of papers at EMNLP 2017. Figures \ref{fig:akhtarExample}, \ref{fig:adelExample} and \ref{fig:guiExample} depict examples of different diagrammatic design choices, including different abstraction levels, content, and representation. 


\section{Requirements}

AI systems orchestrate design choices of \textit{semantic representation}, \textit{algorithms}, \textit{software architecture} and \textit{data resources}. As such, there are multiple levels of abstraction and different use cases to be considered as part of understanding the requirements for a diagrammatic representation. Rather than engage directly with users in the initial stages, we have taken the opportunity to seed this conversation with a formal semiotic slant which will allow for additional thought, formalisation and rigor as the diagrammatic representation evolves. 

\subsection{Use Cases and Target Questions}

In our consideration of the utility of diagrammatic representations for AI systems we adopt two basic personas: That of an \textit{AI Researcher} attempting to review and understand the contribution of the system, and that of an \textit{AI Systems Engineer}, attempting to understand, reproduce or adapt the system architecture. The following list materialise some of the questions which the diagrammatic representation should answer directly. 

\begin{enumerate}
    \item What are the main semantic representation choices in the system? (Relating to Category Theory's \textit{objects} \cite{Eilenberg1945GeneralEquivalences})
    \begin{enumerate}
       \item Latent vs explicit representations, embeddings, data models, compositional models, classification categories, feature types, data integration points.
        \item How the data representation changes along the system?
    \end{enumerate}
    
    \item Which are the main AI algorithmic and functional choices, themes and patterns in the system? (Relating to Category Theory's \textit{morphisms})
    \begin{enumerate}
        \item Which recurring classification categories are used and what is their underlying architecture?
        \item What are the functional composition patterns and their interdependencies?
        \item What are the hyperparameters configurations, and the optimisation ranges? 
    \end{enumerate}
    
    \item What are the main software engineering choices/themes/patterns in the system?
    \begin{enumerate}
        \item What are the temporal dependencies of the components (e.g. asynchronous vs synchronous) and the emerging main architectural themes (e.g. event-based, workflow)?
        \item How are functional units deployed (e.g. in memory, services, or microservices containerised) and how do they communicate?
        \item What are the human interaction points in the system?
    \end{enumerate}
    \item What are the main data management choices/themes/patterns (databases/data model, stream/batch processing) in the system? 
    \item Which performance mechanisms are employed to ensure an efficient data access (query processing, indexing strategies)?
    \item What is the quality delivered by each component and under which evaluation metric? 
    \item What and how are the data resources (knowledge bases, linguistic resources and gold standards) used in the system?
\end{enumerate}

\subsection{DIAL Requirements}
The target questions lead to the following requirements for the representation:

\begin{enumerate}
    \item To target an abstraction-level which is optimal for complex and multi-component AI systems.
    \item To have a primary abstraction level of functional components and data transformation across the system.
    \item To keep a coherent perspective of the whole.
    \item To focus beyond the description of Machine Learning techniques.
    \item To utilise non-linguistic symbolic language for recurring primitives (minimization of verbal elements).
    \item To be extensible to relevant software architecture and data management aspects.
    \item To communicate the performance of each component.
    \item To be cognitively efficient rather than informationally dense.
\end{enumerate}

Quantification, measurement and analysis of the above is intended to take place as DIAL evolves. At this stage, DIAL attempts to satisfy the "representational requirements" but not to completely address all of the "target questions", see Section \ref{section:roadmap}.

\section{Roadmap}
\label{section:roadmap}

In order to allow for different concepts and levels of abstraction, whilst remaining compatible, we are presenting DIAL as a set of dialects. The dialects presented here are first iterations of:
\begin{itemize}
    \item DIAL-SYS, providing the core common language to describe AI Systems at a high level.
    \item DIAL-NN, extending to Neural Network components.
\end{itemize}

The scope of DIAL-SYS is to describe complex (multi-component) AI systems in operation. DIAL-NN extends the DIAL-SYS \textit{symbolset} to facilitate diagrammatic representation of neural network aspects of AI Systems. This area was chosen to illustrate the possibility of dialects, and due to the prevalence of neural networks architectures in current literature. Additional dialects to cover other aspects of AI Systems, such as logic will be published soon. To re-iterate, DIAL, encompassing these dialects, is not presented as a complete framework, but rather a starting point for community dialogue and collaborative construction. It is expected that significant work will be required in order to make DIAL-SYS and DIAL-NN optimal for the AI Community. Once this is in a reasonable shape, and if it proves useful, additional dialects would be expected to evolve in order to tackle different aspects of AI Systems. Future dialects will include:
\begin{itemize}
    \item DIAL-DB, focused on data analysis and data management.  
    \item DIAL-LOG, extending to logic formalisms and systems.
    \item DIAL-ML, specialised for Machine-Learning
    \item DIAL-PGM, for probabilistic graph models such as Bayesian Networks
    \item DIAL-BIZ, to extend to business or application contexts
    \item DIAL-SEM, for semantic knowledge representations \cite{Selby2018ReconstructingPostulates}
\end{itemize}

DIAL-SYS will be developed to incorporate temporal elements, such as execution performance. Extension of DIAL to include representation for "Human-in-the-loop", including external interaction points such as crowd-sourcing for training, will also be required as DIAL matures.

\begin{figure}[htbp]
    \centering
    \includegraphics[scale=0.6]{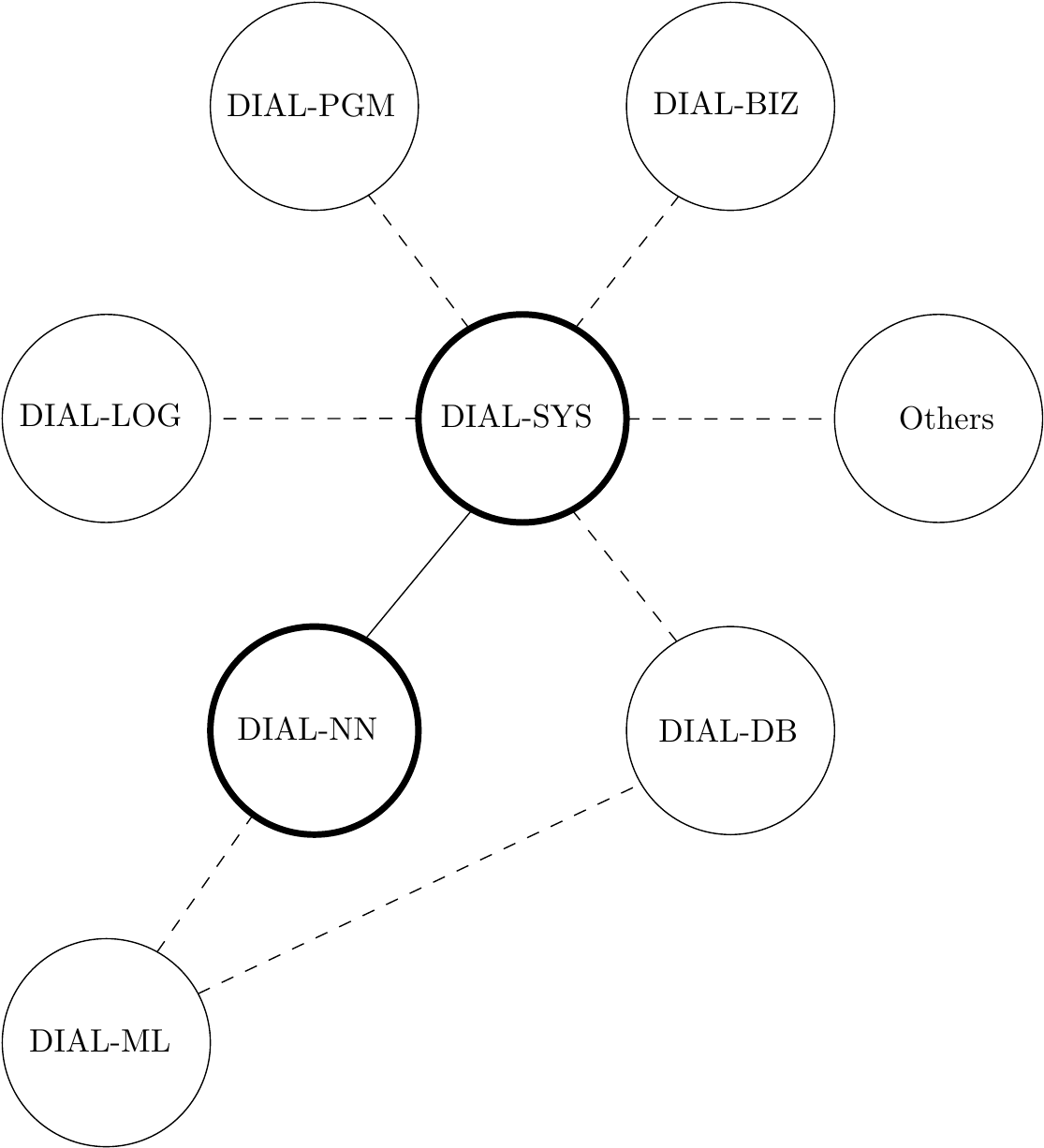}
    \caption{DIAL Architectural Map}
    \label{fig:dialmap}
\end{figure}

Figure \ref{fig:dialmap} contains an incomplete view of some of the expected DIAL dialects (DIAL-ects). Dark circles show existing work, and lines between the circles indicate significant dependency between dialects. The prioritisation and timeline will be based on the comments received, and led by the future stewards of this standardisation effort who it is hoped will come forward (see Section \ref{section:rfc}). 

\section{Applications}

With a more defined scope for DIAL, we continue to detail some of the anticipated applications of DIAL. The following subsections explore some of the possibilities of DIAL to support AI community members.

\subsection{Nano-publications}

With such a high pace of change and volume of conference papers in AI, it is extremely difficult to keep on top of the latest developments in the field. One nano-publication, consisting of a DIAL diagram, can summarise the composition of the functional elements of the proposed model, their associated performance and core architectural decisions as well as the contribution. 

\subsection{Software specification}

DIAL, as a language, allows for specification of software by prioritising recurring AI components in the representation. This focus on AI by design can provide a more  concise representation, focusing on the articulation of the AI-related design decisions (in contrast with UML).  

\subsection{Designing AI Systems}
There is the potential, with substantial development effort, to extend DIAL to be able to simulate AI Systems in a way analogous to Electronic Design Automation for chips. This could allow for rapid prototyping and facilitate reasoning about novel architectures. Differently from TensorBoard, this would be created to support multiple different dataflow libraries, and aim to be technology agnostic. 

\subsection{Estimating the Impact of AI Systems in Businesses}
DIAL can serve as a lower entry barrier for communicating AI functionalities and their associated performance levels to non-AI experts stakeholders. This can serve as a mechanism to communicate requirements and to synchronize expectations between business and technical profiles. This expected that DIAL can improve the formalisation of the requirements (with a lower effort) and the quality of AI software estimation.

\subsection{Education}
In education, the creation of curricula describing the different perspectives around the construction of AI systems is still an open challenge. DIAL can reduce the barriers for conceptualising AI systems under a complex/multi-technique perspective, providing an integrative view of AI. Moreover, currently there is a substantial gap between the discussions within the research community and their industrial application. By reducing the interpretation bottleneck of existing scientific contributions, it is expected that DIAL can serve as a fundamental educational device for AI practitioners and researchers, catalyzing experimentation across different communities. 


\section{DIAL-SYS Components}
\label{section:components}

For interpreting DIAL it is assumed that users will have familiarity with elements of AI Systems, and as such there is an assumption that common acronyms are understood and need not be specified. As an example, "Bi-directional long-short term memory architecture" is written as BiLSTM without special introduction. Many components and symbols do require more definition: Table \ref{table:notationalindex} contains an index to DIAL notation conventions.

\begin{table}[htbp]
    \caption{DIAL notational index}
    \label{table:notationalindex}
    \vskip 0.15in
\begin{center}
\begin{small}
\begin{sc}
    \centering
    \begin{tabular}{lr}
     \toprule
Table Number & Content and associated conventions\\
\midrule
~\ref{table:bigtable} & AI shapes, domain and range.\\
~\ref{tab:operatorsSYS} & DIAL-SYS symbolic elements.\\
~\ref{table:datatypes} & Data Types.\\
~\ref{table:exampleclasses} & Some example classifications.\\
~\ref{tab:operatorsNN} & DIAL-NN symbolic elements.\\ 
\bottomrule
    \end{tabular}
\end{sc}
\end{small}
\end{center}
\vskip -0.1in 
\end{table}

\subsection{Symbols}

The iconic representation of DIAL, with its semiotic properties, provides the structuring framework for the diagrammatic representation. The DIAL-SYS \textit{symbolset} provides a representation for high-level AI system components (as depicted in Table ~\ref{table:bigtable}).

\begin{table*}[htbp]
    \caption{Shape inputs and outputs}
    \label{table:bigtable}
    \vskip 0.15in
\begin{center}
\begin{small}
\begin{sc}

    \centering
    \begin{tabular}{p{7cm}p{4cm}p{2.5cm}}
     \toprule
    Shape & Domain & Range \\
   \midrule
POS Tagging (POS) & $S$ & $S^{POS}$\\
Syntactic Parsing (SYN) & $S^{(POS)}$ & $PN$ Structure\\
\midrule
Named Entity Recognition (NER) & $S^{(Chunk)}$ & $S^{Names}$\\
Word Sense Disambiguation (WSD) & $S^{POS,Chunk}$, $KB$ & $Term^{WSD}$ or wordnet IDs\\
Entity Linking (EL) & $S^{NER}$ & Entity,(updated $KB$)\\
Semantic Role Labeling (SRL) & $S^{Token}$ & $S^{Sem}$\\
Semantic Relation Classification (SRC) & $Term_1$,$Term_2$ & $Term^{Sem}$\\
Open Relation Extraction (OIE)& $S$ & ${Pred(Arg)}$\\
Predicate Creation & $(T)(Data)$ & $Term_{New}$ + definition\\
\midrule
Structured Data Querying & $?q, KB$ & Tuples with labels\\
Text Retrieval & human $T$ & human-readable $T$?\\
\midrule
Natural Language Generation & $Pred(Arg)$ & human-readable $T$\\
Text Simplification & $T$ & $T$\\
Text Summarisation & $T^{(Chunk)(NER)(Arg)(Name)(Sem)}$ & $T$\\
\midrule
Co-reference resolution (COREF) & $S^{NER}$ or $T^{(Token)}$ & \{Chains with IDs\}\\
Rhetorical structure classification (RST) & $(S)$;($S_1,S_2$);$(T)$ &  $S^{RS}$;$T^{RS}$\\
Argumentation structure classification & $[s_1,s_2],T$ & $T^{ArgStruct}$\\
Argument Scheme classification & $T^{ArgStruct}$ & $C^{ArgScheme}$\\
\midrule
Polarity and emotion analysis & $S^{(Pred(Arg))}$ & Score\\
Rhetorical figures analysis & $T$ & $T^{RST}$\\
\midrule
String similarity & $string_1,string_2$ & Score\\
Semantic similarity & $terms^{(entity)}$ & Score\\
Semantic relatedness & $terms^{(entity)}$ & Score \\
\midrule
Inductive reasoning & $Pred(Arg)^{F}$, ($KB_R$), $KB_{Constraints}$ & $S^{(Pred(Arg))}$\\
Deductive reasoning & $Pred(Arg)$+$KB_{F,R}$ & $Pred(Arg)$\\
Abductive reasoning & $Pred(Arg)^{F}$, $KB$ & $Pred(Arg)$ sequence\\
\bottomrule
    \end{tabular}
    
\end{sc}
\end{small}
\end{center}
\vskip -0.1in 
\end{table*}

\begin{table}[htbp]
    \caption{DIAL-SYS Symbolic Elements}
    \label{tab:operatorsSYS}
    \vskip 0.15in
    \begin{center}
    \begin{small}
    \begin{sc}
    \begin{tabular}{lc}
        \toprule
        Function & Symbol \\
        \midrule
        Direct sum & $\bigoplus$ \\
        Concatenation & $+\!\!+$ \\ 
        Tensor product & $\bigotimes$ \\      
  
        Set & $\{elements\}$\\
        Data flow & $\rightarrow$ \\
        Data flow (both ways) & $\leftrightarrow$ \\
        KB Query & \includegraphics[scale=0.5]{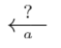}\\
        Data persistence & $\longmapsto$ \\
        Conditional & \includegraphics[scale=0.6]{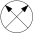} \\
        System interface (e.g. service, API) & $\multimap$ \\
        Composition & a $\circ$ b \\
        Join & $\bowtie$ \\
        Similarity \& Relatedness & $\measuredangle \theta$ \\
        Embedding projection (with identifier) & $\vec{\Pi}_{id}$ \\
        Word2vec & $w2v$ \\
        Similarity \& Relatedness (if not cosine, specify) & $\measuredangle \theta$ \\
        Regression & \includegraphics[scale=0.5]{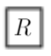} \\
        Classifier & \includegraphics[scale=0.5]{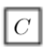} \\
        Classification & $C$ \\
        Ranking operator & $R\uparrow$\\
        Top n elements & $R\uparrow n$\\ 
        Encoder & \includegraphics[scale=0.5]{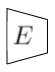}\\
        Decoder & \includegraphics[scale=0.5]{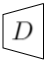}\\
        Deductive Reasoning & $E \vDash$ \\
        Verification (e.g. user validation) & $\CheckedBox$ \\
        Function & \includegraphics[scale=0.5]{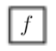} \\
        Function (contraction) &  \includegraphics[scale=0.5]{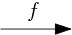}  \\
        Datasets, Data resources & \includegraphics[scale=0.4]{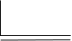} \\
        Gold Standard & \includegraphics[scale=0.6]{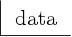} \\
        Knowledge Base of functions & \includegraphics[scale=0.4]{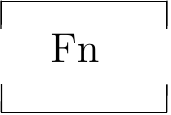}\\
        Zoom in & Dotted line with box \\ 
        Accuracy & $acc$\\ 
        \bottomrule
   \end{tabular}
\end{sc}
\end{small}
\end{center}
\vskip -0.1in
\end{table}

The primary DIAL-SYS \textit{symbolset} can be extended with more specific symbolsets. Section ~\ref{section:NN} provides a description of DIAL-NN, a dialect aiming towards the standardisation of NN architectures.

\subsection{Data Types \& Notation Convention} \label{section:diagraminfo}

DIAL symbols are typed with subscripts and superscripts, which describe recurring \textit{classification tasks}, their associated \textit{classes} or \textit{data types}, tensor dimensionality, common data models (file formats). Table \ref{table:notationalindex} indexes the tables. Features which form parts of the architecture are in \textit{circles}, and full components are in \textit{rectangles}. In this way, it is possible to see at a glance the level at which the system is abstracted.


\begin{table*}[htbp]
    \caption{Data Types}
    \label{table:datatypes}
    \vskip 0.15in
\begin{center}
\begin{small}
\begin{sc}
    \centering
    \begin{tabular}{p{2cm}p{1.2cm}p{7cm}}
     \toprule
Category & Notation & Description \\
\midrule
Text & $T$ & Raw file.\\
Passage & $p_T$ & Fragment of text.\\
Sentence & $s_T$ & Set of words from a text, that is complete in itself.\\
Character & $Ch_T$ & A single printable symbol in a text.\\
Term & $t_T$ & A word or set of words in a text that express a concept.\\
Word & $w_T$ & A single distinct meaningful element of text.\\
Dialogue term & $dt$ & A set of words forming a distint part of a dialogue.\\
Sense & $\overline{w_T}$ & A specifically disambiguated word, including ID.\\
Clustered word & $\dot{w_T}$ & Word embedded in a Vector Space.\\
Image & $im$ & Raw file.\\
Sentence Sense & $\overline{s}_T$ & An identified sentence, distinguished from minor or adjacent sentences.\\
Query & $?q$ & A set of words as a query input. \\
Answer & $\hat{a}$ & A set of words as an answer output. \\
Facts & $F$ & A predicate with an n-tuple of word constants, satisfied unconditionally.\\ 
Rules & $R$ & A statement which gives conditions under which tuples of words satisfy a predicate. \\
Classification outcome & $P_c[a,b]$ & The probability distribution of a classification $c$, with range $[a,b]$.\\ 
\bottomrule
    \end{tabular}
\end{sc}
\end{small}
\end{center}
\vskip -0.1in 
\end{table*}

\begin{table}[htbp]
    \caption{Example Classifications}
    \label{table:exampleclasses}
    \vskip 0.15in
\begin{center}
\begin{small}
\begin{sc}
    \centering
    \begin{tabular}{lcr}
     \toprule
Category & Notation \\
\midrule
NER classified sentence & $S^{NER}$ \\
SRL classified sentence & $S^{SRL}$ \\
POS classified sentence & $S^{POS}$ \\
Argument Scheme Classification & $C^{ArgScheme}$ \\
Argumentation Structure & $T^{ArgStruct}$ \\
Classified ambiguous terms & $Term^{WSD}$ \\
$Pred(Arg)$ labeled with Facts & $Pred(Arg)^{F}$ \\
\bottomrule
    \end{tabular}
\end{sc}
\end{small}
\end{center}
\vskip -0.1in 
\end{table}

\section{DIAL-NN}
\label{section:NN}
As a second dialect, DIAL-NN (Neural Networks) was prioritised due to their prevalence in recent literature. As mentioned in the survey of Section \ref{section:heterogeneity}, 77\% of the papers surveyed were designing new Neural Networks or applying NN-based architectures.

\subsection{Symbols}
DIAL-NN symbolic elements differ from DIAL-SYS symbolic elements due to distinguishing semiotic requirements. Table \ref{tab:operatorsNN} shows the symbolic elements of this dialect.

\begin{table}[htbp]
    \caption{DIAL-NN Symbolic Elements}
    \vskip 0.15in
    \begin{center}
    \begin{small}
    \begin{sc}
    \begin{tabular}{lc}
        \toprule
        Function & Symbol \\
        \midrule
        Loss function & $\Delta$\\
       Activation function (label with e.g. tanh) & \includegraphics[scale=0.3]{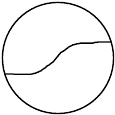}\\
        Softmax & \includegraphics[scale=0.3]{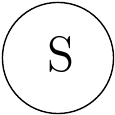}\\
        Attention & \includegraphics[scale=0.3]{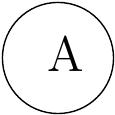}\\
        RNN Layer (eg LSTM) & \includegraphics[scale=0.2]{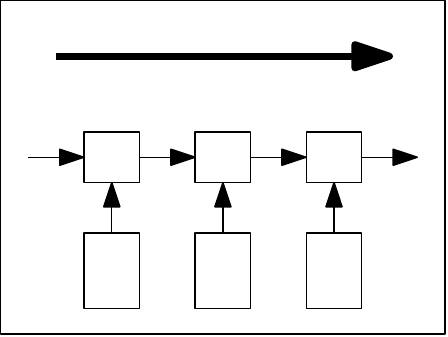}\\ 
        BiLSTM Layer & \includegraphics[scale=0.2]{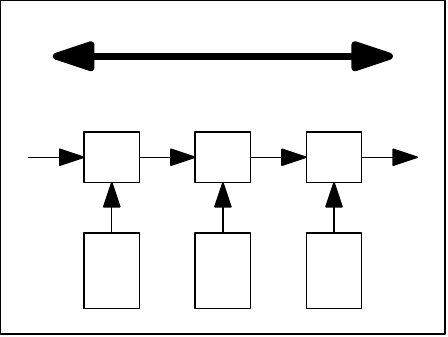}\\ 
        GRU Layer &  \includegraphics[scale=0.2]{viktorlstm-eps-converted-to}GRU \\ 
        Convolutional Layer & \includegraphics[scale=0.15]{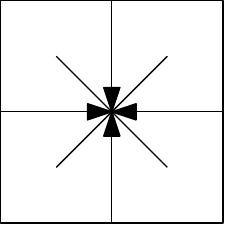} \\
        Recursive Neural Network & \includegraphics[scale=0.15]{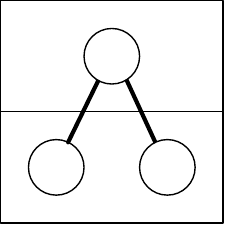}\\
        Support Vector Machine & \includegraphics[scale=0.1]{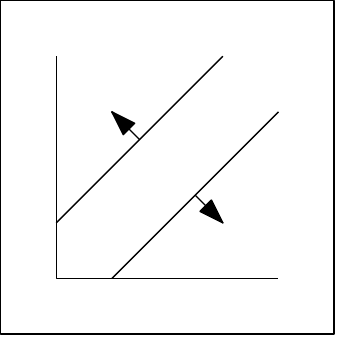}\\\\
        Ground truth of sentiment classification & $g_c$ \\
        Hidden layer (forward directionality)& $\overrightarrow{h}$\\
        Hidden layer (backward directionality)& $\overleftarrow{h}$\\
        \ldots & \ldots \\
        \bottomrule
   \end{tabular}
\end{sc}
\end{small}
\end{center}
    \label{tab:operatorsNN}
\vskip -0.1in
\end{table}

\clearpage
\section{DIAL by Example}

\subsection{Example 1: QA System}
\begin{figure}[htbp]
    \centering
    \includegraphics[scale=0.6]{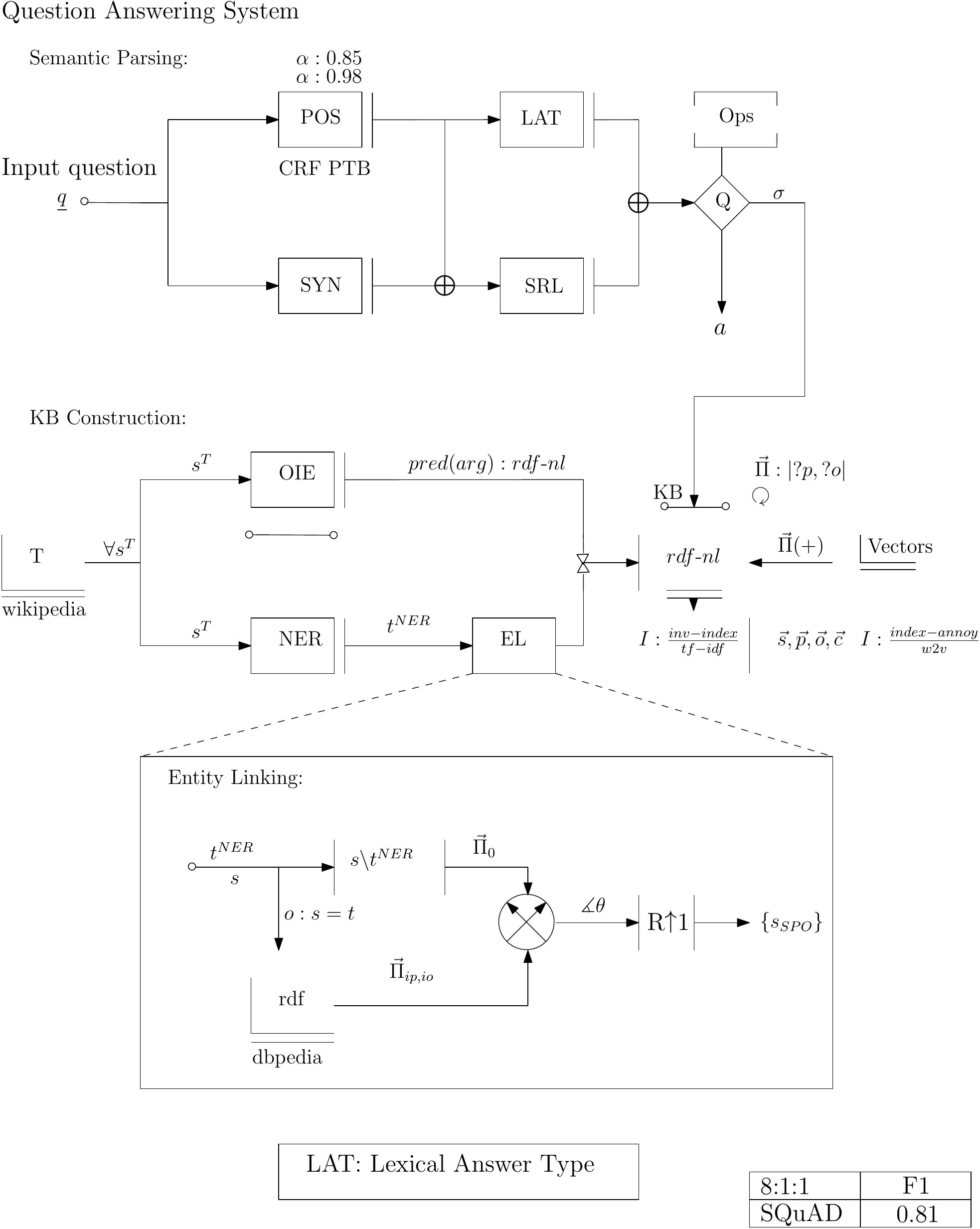}
    \caption{DIAL Diagram for a hypothetical QA system}
    \label{fig:QAdiag}
\end{figure}

Figure \ref{fig:QAdiag} depicts an example DIAL-SYS diagram for a question answering system over unstructured text. The diagram depicts two cycles: (i) \textit{KB construction}, which transforms textual data into a structured KB and (ii) \textit{semantic parsing}, which transforms a natural language query into a formal structured query representation over the KB extracted in (i). 

In the \textit{KB construction} cycle, for each document $d$ the pipeline iterates through all the sentences performing \textit{open information extraction} (OIE) (as an external service),  \textit{named entity recognition} (NER) and \textit{entity linking} (EL). This diagram does not describe in this context the OIE and NER steps, however it details the EL step. The EL starts by taking the complement to the NER chunks within the context of the sentence. In parallel, it collects the entity types associated to a structured KB (DBpedia). Both result sets are projected into a pre-built w2v space, where a cross-product between both sets is performed, cosine similarity measures are calculated and the top 1 (\textit{argmax()}) is calculated. The output of the OIE component is joined to the output of the NER and EL, which is serialised into a RDF-NL file and indexed using an inverted index using the  \textit{tf-idf weighting scheme}. The RDF-NL file is composed with a pre-built word embedding, and this KB is exposed as a service. 

In the \textit{semantic parsing} cycle, a natural language question is sent to two parallel chains. In the upper part of the diagram, a POS tagger is applied, where the classifier type and the associated corpus are specified, as well as the accuracy in the original corpus and the accuracy in the current domain, this is then sent to a  \textit{lexical answer type} (LAT) detection component. In the lower part of the diagram, a C-Structure syntactic parsing is performed. Its output is added to the output of the POS tagger and sent to the SRL tagger. The LAT and the SRL are sent to a  \textit{Q-learning component}, which uses the KB of actions \textit{Op} (queries over the structured KB - not detailed in this example) and the LAT + SRL, to learn a sequence of operations $\sigma$. The final output is given by the answer $a$.

\clearpage
\subsection{Example 2: Lexicon-based Attention}
\begin{figure}[htbp]
    \centering
    \includegraphics[scale=0.55]{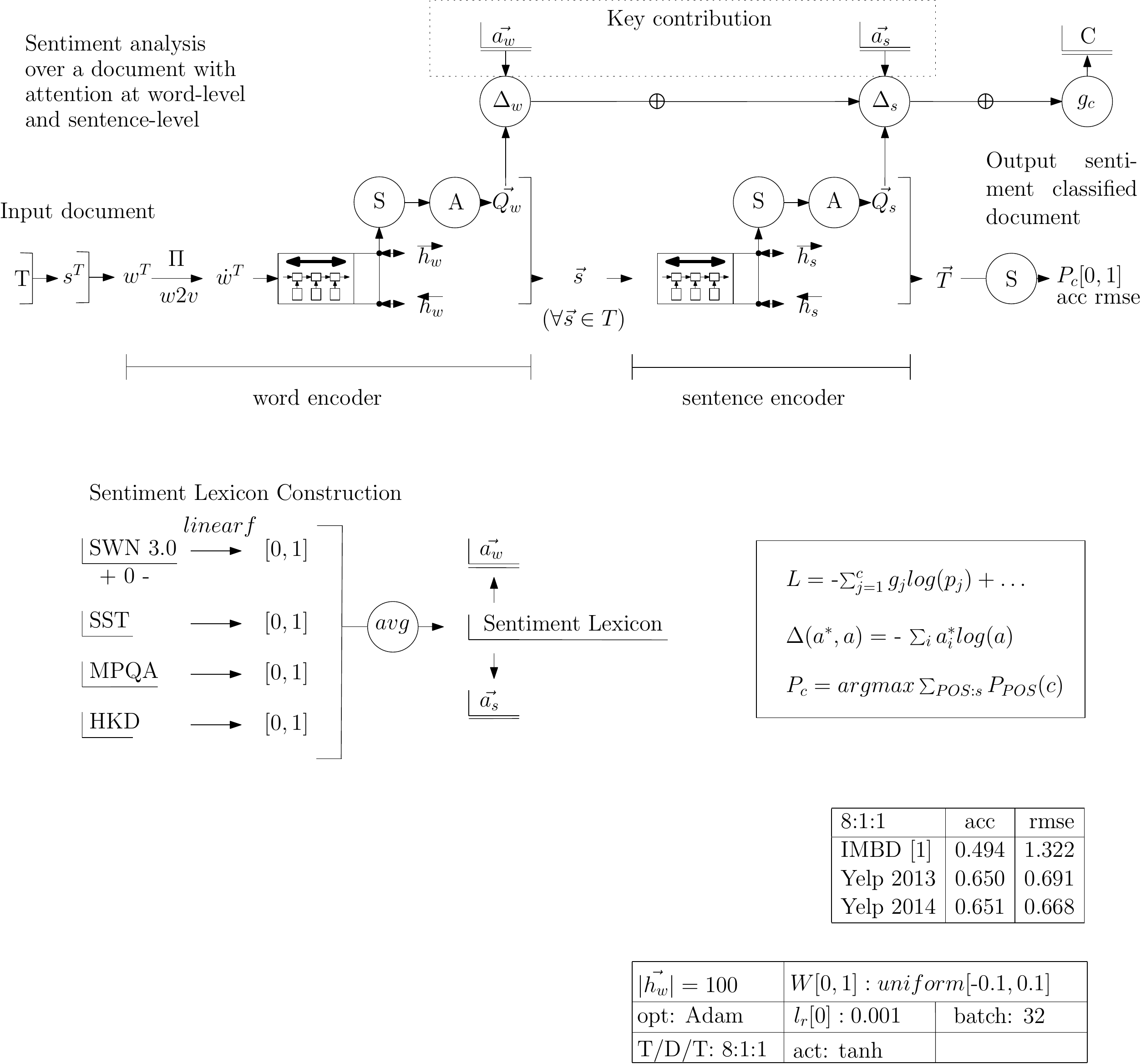}
    \caption{DIAL Diagram for Zou et al's Lexicon-based supervision attention model}
    \label{fig:NNdiag}
\end{figure}

The DIAL-NN dialect is illustrated by depicting the model described in Zou et al \cite{Zou2018AAnalysis}, which performs sentiment analysis over a document with attention at word-level and at the sentence level. Figure \ref{fig:NNdiag} depicts the organization of the architecture where the up most diagram describes the two Bi-LSTM layers with attention at the word-level and at the sentence level. Both models are jointly trained based on the creation of a word-level and sentence-level attention lexicon (lower left part). The hyper-parameters and the model accuracy are described in the two tables.

\newpage
\subsection{Example 3: Text Entailment}
\begin{figure}[htbp]
    \centering
    \includegraphics[scale=0.6]{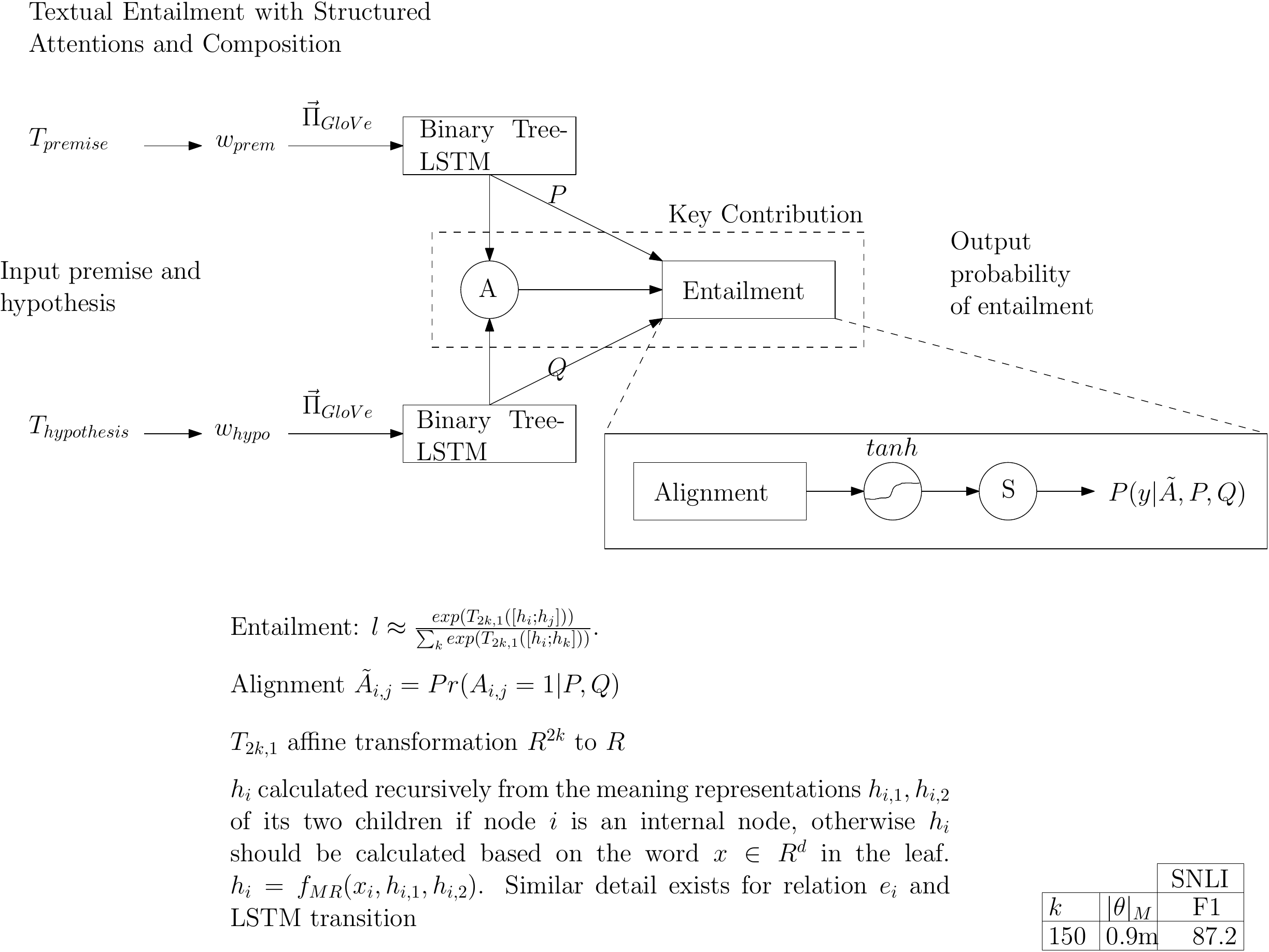}
    \caption{DIAL Diagram for Textual Entailment with Structured Attentions and Composition \cite{Zhao2016TextualComposition}}
    \label{fig:entailment}
\end{figure}

Figure \ref{fig:entailment} is a DIAL-SYS for Zhao et al \cite{Zhao2016TextualComposition}. Note that the details of the Neural Network are not included in this DIAL, though it could be extended to show this. Two Siamese projections (with shared weights) occur for premise and hypotheses, in order to create binary-tree LSTMs. A dual-attention model is applied to tree nodes between premise and hypothesis. The attention and outputs from binary trees for both premise and hypothesis are used in the entailment. At each hypothesis node i, $e_i$ is calculated recursively given the meaning representation at this tree node $h_i$, the meaning representation of every node in the premise tree $h_j$ , $j \in P$, and the entailment from i’s children, $e_{i,1}, e_{i,2}$. The entailment alignment is used in a tanh activation function for softmax, in order to output the probability of entailment between a premise and hypothesis. Note that this is a simplification due to forget gates etc.

\section{Creating DIAL Diagrams}

\subsection{Practicalities}
The DIAL examples contained in this report were made with IPE, a LaTeX compatible graphics editor (download source \url{http://ipe.otfried.org}). An accompanying DIAL resource library will be created in due course. 

\subsection{Semiotic Principles}

By "semiotics", we follow Morris in his definition of the field as encompassing semantics (what is represented), syntactics (relationships between things), and pragmatics (their interpretation within context) \cite{Morris1938FoundationsSigns}.

Spatial arrangement of symbols, arrows and natural language (Section \ref{section:components}) attempts to be reasonably standardised but not overly prescriptive. Additionally, they inherit existing common practices collected from the observation of the literature (see Survey section ??).

Top left title, bottom right data and hyperparameters are example of macroscopic guidelines. On a more detailed level, left-to-right linear is the default with exceptions where it makes sense (such as Siamese Networks, see Figure \ref{fig:entailment}). Arbitrary alignment or misalignment is not encouraged, as they can confuse the representation. Making use of "intuitive" spatial patterns is also advisable, such as if the input is on the left, keep the input on the left in a "zoom in". Any necessary functions, if necessary to detail, should be included in the mathematical key with notation consistent with the accompanying paper.

\begin{figure}[htbp]
    \centering
    \includegraphics[scale=0.7]{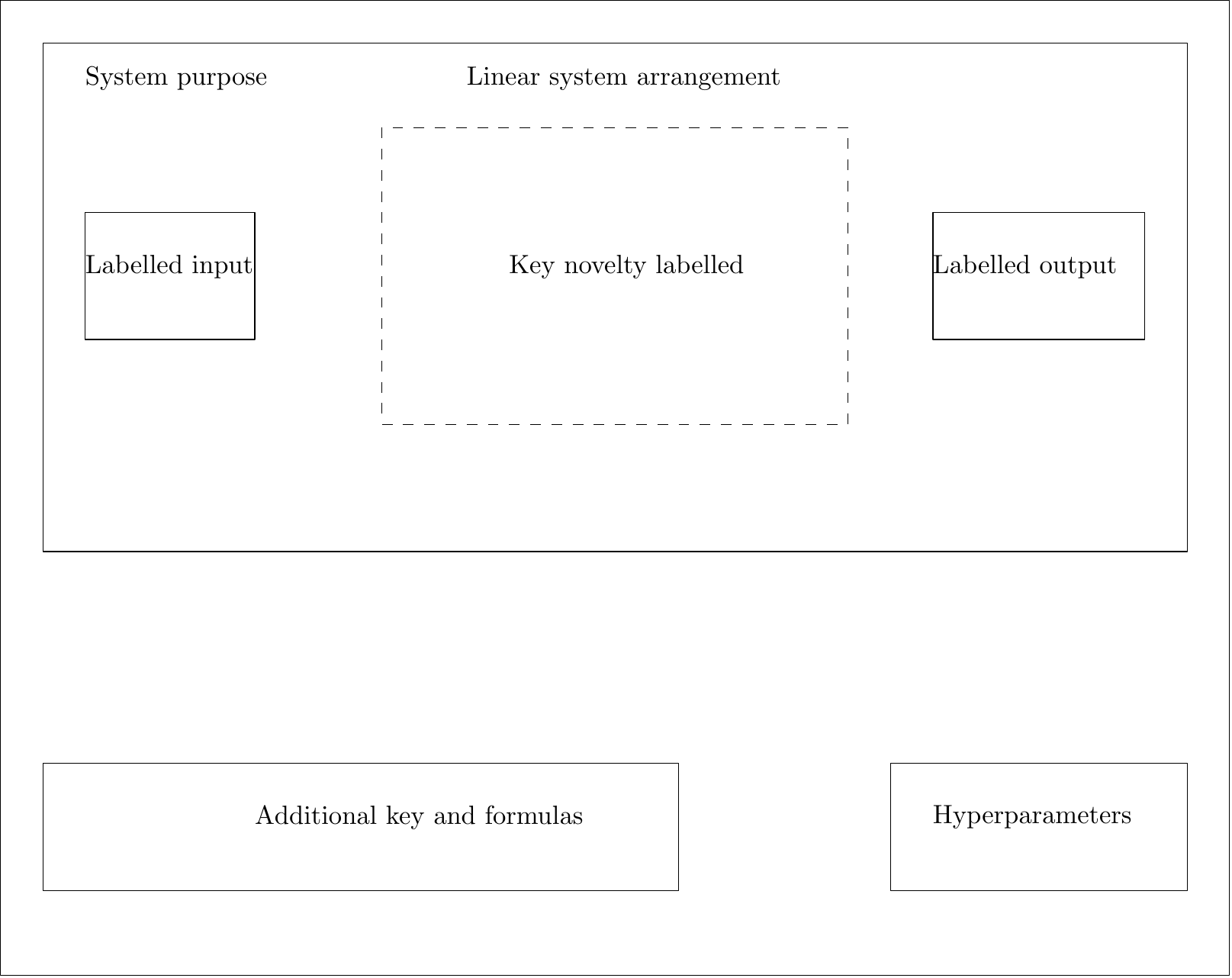}
    \caption{DIAL Schematic Outline}
    \label{fig:schema}
\end{figure}

Introduce new symbols and relational concepts sparingly: We appreciate that DIAL is a work in progress and that some useful symbols will be missing. For example, it may be tempting to create a new symbol for linear scaling functions: Please avoid doing this, instead using existing natural language or mathematical notation. We are also aware that the practice of embedding text within an image is not ideal, and that there is a need for DIAL-NN to incorporate more low-level architectures (along the lines of Graves et al's diagrams \cite{Graves2013HybridLSTM}). With interested community members we will be looking to adapt and extend DIAL in order to aid its usefulness. Please see Section \ref{section:rfc} on how to become a contributor of DIAL.

\section{Request for Comments (RfC)}
\label{section:rfc}

Particularly at this early stage of development, feedback and interaction is sought from AI practitioners of all types. Feedback is especially helpful on topics that will aid completeness, consistency and usability of the existing scope of DIAL-SYS and DIAL-NN. As a longer-term commitment, those interested in being stewards for the standardisation effort should also get in touch with the authors. Further information, including contact details, can be found at \url{aidiagrams.com}.

\section*{Acknowledgements}
The authors would like to thank Viktor Schlegel, Chaohui Marshall and three anonymous reviewers for their useful comments. 

\bibliographystyle{acm}
\bibliography{main.bbl}

\begin{thebibliography}{10}

\bibitem{Adel2017GlobalClassification}
{\sc Adel, H., and Sch{\"{u}}tze, H.}
\newblock {Global Normalization of Convolutional Neural Networks for Joint
  Entity and Relation Classification}.
\newblock In {\em Proceedings of the 2017 Conference on Empirical Methods in
  Natural Language Processing\/} (2017), Association for Computational
  Linguistics, p.~1723–1729.

\bibitem{Akhtar2017AAnalysis}
{\sc Akhtar, S., Kumar, A., Ghosal, D., Ekbal, A., and Bhattacharyya, P.}
\newblock {A Multilayer Perceptron based Ensemble Technique for Fine-grained
  Financial Sentiment Analysis}.
\newblock In {\em Proceedings of the 2017 Conference on Empirical Methods in
  Natural Language Processing\/} (2017), pp.~540--546.

\bibitem{Booch1998UnifiedThe}
{\sc Booch, G., Rumbaugh, J., Jacobson, I., and Wesley, A.}
\newblock {\em {Unified Modeling Language User Guide, The}}.
\newblock Addison Wesley, 1998.

\bibitem{Eilenberg1945GeneralEquivalences}
{\sc Eilenberg, S., and MacLane, S.}
\newblock {General theory of natural equivalences}.
\newblock {\em Transactions of the American Mathematical Society 58\/} (1945),
  231--231.

\bibitem{Glinz2000ProblemsLanguage}
{\sc Glinz, M.}
\newblock {Problems and Deficiencies of UML as a Requirements Specification
  Language}.
\newblock In {\em Proceedings of the 10th International Workshop on Software
  Specification and Design\/} (2000), Institut f{\"{u}}r Informatik,
  Universit{\"{a}}t Z{\"{u}}rich, IEEE Computer Society, pp.~11--22.

\bibitem{Graves2013HybridLSTM}
{\sc Graves, A., Jaitly, N., and Mohamed, A.-R.}
\newblock {Hybrid Speech Recognition with Deep Bidirectional LSTM}.
\newblock In {\em Automatic Speech Recognition and Understanding (ASRU), 2013
  IEEE Workshop on\/} (2013), IEEE, pp.~273--278.

\bibitem{Gui2017Part-of-SpeechNetworks}
{\sc Gui, T., Zhang, Q., Huang, H., Peng, M., and Huang, X.}
\newblock {Part-of-Speech Tagging for Twitter with Adversarial Neural
  Networks}.
\newblock In {\em Proceedings of the 2017 Conference on Empirical Methods in
  Natural Language Processing\/} (2017), pp.~2411--2420.

\bibitem{Harrison2005TheGlobalization}
{\sc Harrison, N.}
\newblock {The Darwin information typing architecture (DITA): applications for
  globalization}.
\newblock In {\em IPCC 2005. Proceedings. International Professional
  Communication Conference, 2005.\/} (2005), IEEE, pp.~115--121.

\bibitem{Moody2007WhatDevelopment}
{\sc Moody, D.}
\newblock {What Makes a Good Diagram? Improving the Cognitive Effectiveness of
  Diagrams in IS Development}.
\newblock In {\em Advances in Information Systems Development}. Springer US,
  Boston, MA, 2007, pp.~481--492.

\bibitem{Morris1938FoundationsSigns}
{\sc Morris, C.~W.}
\newblock {Foundations of the theory of signs}.
\newblock {\em The Journal of Symbolic Logic 3}, 04 (12 1938), 158.

\bibitem{Otero2005AnUML}
{\sc Otero, M.~C., and Dolado, J.~J.}
\newblock {An empirical comparison of the dynamic modeling in OML and UML}.
\newblock {\em Journal of Systems and Software 77}, 2 (8 2005), 91--102.

\bibitem{Selby2018ReconstructingPostulates}
{\sc Selby, J.~H., Scandolo, C.~M., and Coecke, B.}
\newblock {Reconstructing quantum theory from diagrammatic postulates}, 2018.

\bibitem{Sethi2018DLPaper2Code:Papers}
{\sc Sethi, A., Sankaran, A., Panwar, N., Khare, S., and Mani, S.}
\newblock {DLPaper2Code: Auto-generation of Code from Deep Learning Research
  Papers}.
\newblock In {\em Thirty-Second AAAI Conference on Artificial Intelligence\/}
  (2018), IBM Research, pp.~7339--7346.

\bibitem{Shahin2014ATechniques}
{\sc Shahin, M., Liang, P., and Babar, M.~A.}
\newblock {A systematic review of software architecture visualization
  techniques}.
\newblock {\em The Journal of Systems and Software 94\/} (2014), 161--185.

\bibitem{Sloman1984WhyFormalisms}
{\sc Sloman, A.}
\newblock {Why We Need Many Knowledge Representation Formalisms}.
\newblock In {\em Proceedings BCS Expert Systems\/} (1984), University of
  Sussex, Brighton, UK, Cambridge University Press, pp.~163--183.

\bibitem{Zhao2016TextualComposition}
{\sc Zhao, K., Huang, L., and Ma, M.}
\newblock {Textual Entailment with Structured Attentions and Composition}.
\newblock In {\em Proceedings of COLING 2016, the 26th International Conference
  on Computational Linguistics: Technical Papers\/} (2016), pp.~2248--2258.

\bibitem{Zou2018AAnalysis}
{\sc Zou, Y., Gui, T., Zhang, Q., and Huang, X.}
\newblock {A Lexicon-Based Supervised Attention Model for Neural Sentiment
  Analysis}.
\newblock In {\em Proceedings of the 27th International Conference on
  Computational Linguistics\/} (2018), pp.~868--877.

\end{thebibliography}

\end{document}